# Autoencoder-Based Detection of Anomalous Stokes V Spectra in the Flare-Producing Active Region 13663 Using Hinode/SP Observations

Jargalmaa BATMUNKH[*1], Yusuke IIDA[*1], Takayoshi OBA[*2]

## ABSTRACT

Detecting unusual signals in observational solar spectra is crucial for understanding the features associated with impactful solar events, such as solar flares. However, existing spectral analysis techniques face challenges, particularly when relying on pre-defined, physics-based calculations to process large volumes of noisy and complex observational data. To address these limitations, we applied deep learning to detect anomalies in the Stokes V spectra from the Hinode/SP instrument. Specifically, we developed an autoencoder model for spectral compression, which serves as an anomaly detection method. Our model effectively identifies anomalous spectra within spectro-polarimetric maps captured prior to the onset of the X1.3 flare on May 5, 2024, in NOAA AR 13663. These atypical spectral points exhibit highly complex profiles and spatially align with polarity inversion lines in magnetogram images, indicating their potential as sites of magnetic energy storage and possible triggers for flares. Notably, the detected anomalies are highly localized, making them particularly challenging to identify in magnetogram images using current manual methods.

## 1. INTRODUCTION

Solar spectra provide physical parameters such as Doppler velocity, temperature, and magnetic fields. Identifying unusual signals in the spectra is significantly important from the perspective of observational discovery. The Hinode Solar Optical Telescope Spectral-Polarimeter (Hinode SOT/SP) mission [1,2] has continuously collected high-resolution, full-Stokes solar spectro-polarimetric data [3]. However, conventional methods for analyzing these spectral data lack flexibility because they rely on time-consuming, human-driven setups for statistical calculations. Additionally, high noise levels in the observations complicate data processing and often necessitate manual inspections. Furthermore, the rapidly increasing volume of observational data collected by the satellite further challenges the ability of conventional methods to examine each spectrum individually.

Machine learning (ML) approaches have demonstrated increasing efficacy in handling large-scale, complex data, making them well-suited for processing and analyzing solar spectral data. Previous studies on solar flare prediction using spectral data [4,5] have applied ML techniques primarily focused on Stokes I profiles, which contain information mainly on temperature and velocity, with pre-selected features for model input. Incorporating the other Stokes profiles (Q, U, and V) would enhance the analysis to improve flare prediction accuracy because these profiles carry intrinsic information strongly influenced by the magnetic field in the solar atmosphere. To address this, we propose a method specifically adapted to enable detecting anomalous profiles in both Stokes I and V spectra from Hinode/SP data, using a compression

[*1] Niigata University
[*2] Max Planck Institute for Solar System Research

model [6] based on a 1D-convolutional autoencoder architecture.

Because X-class flares are known to have particularly significant impacts on the Earth, we tested our method on recent data from the X1.3 flare observed in active region (AR) 13663 in May 2024. Our analysis centered on reconstructing Stokes V profiles due to their high magnetic sensitivity for examining the spatial distributions of anomalous Stokes V spectra in correlation with magnetogram images.

## 2. METHOD

### 2.1. Autoencoder

#### 2.1.1. Model architecture

In deep learning, deep autoencoders are a powerful model architecture designed for non-linear compression tasks. Their core principle involves compressing data by reconstructing the data through two main components: the encoder and the decoder. The encoder reduces the dimensionality of the input into a compact feature vector, while the decoder expands this feature vector back to its original dimensions.

To facilitate dimensional reduction and reconstruction, both the encoder and decoder utilize various neural network layers, such as fully connected and convolutional layers. Our model uses 1D-convolutional and deconvolutional layers, complemented by pooling and upsampling layers, to compress and reconstruct Stokes I and V spectra efficiently. The architecture of the autoencoder model is detailed in Figure 1.

#### 2.1.2. Autoencoder as an anomaly detector

Due to their efficient compression capabilities, autoencoders are used as anomaly detectors by learning to reconstruct normal data and by identifying deviations in reconstruction for anomalous data. The model is initially trained on a normal dataset containing no anomalies, allowing it to learn the underlying patterns in the data. As a result, the autoencoder can reconstruct any new normal data with minimal reconstruction error. However, when the pre-trained model encounters anomalous data that it has not seen before, it struggles to achieve accurate reconstruction, leading to a substantially higher reconstruction error. By analyzing this error, we can distinguish anomalous data from normal data. In our study, we apply this principle using our compression model to detect unusual spectral profiles in solar spectral data prior to flare events and explore whether or not these spectra correlate with actual flare-triggering regions in the spatial domain.

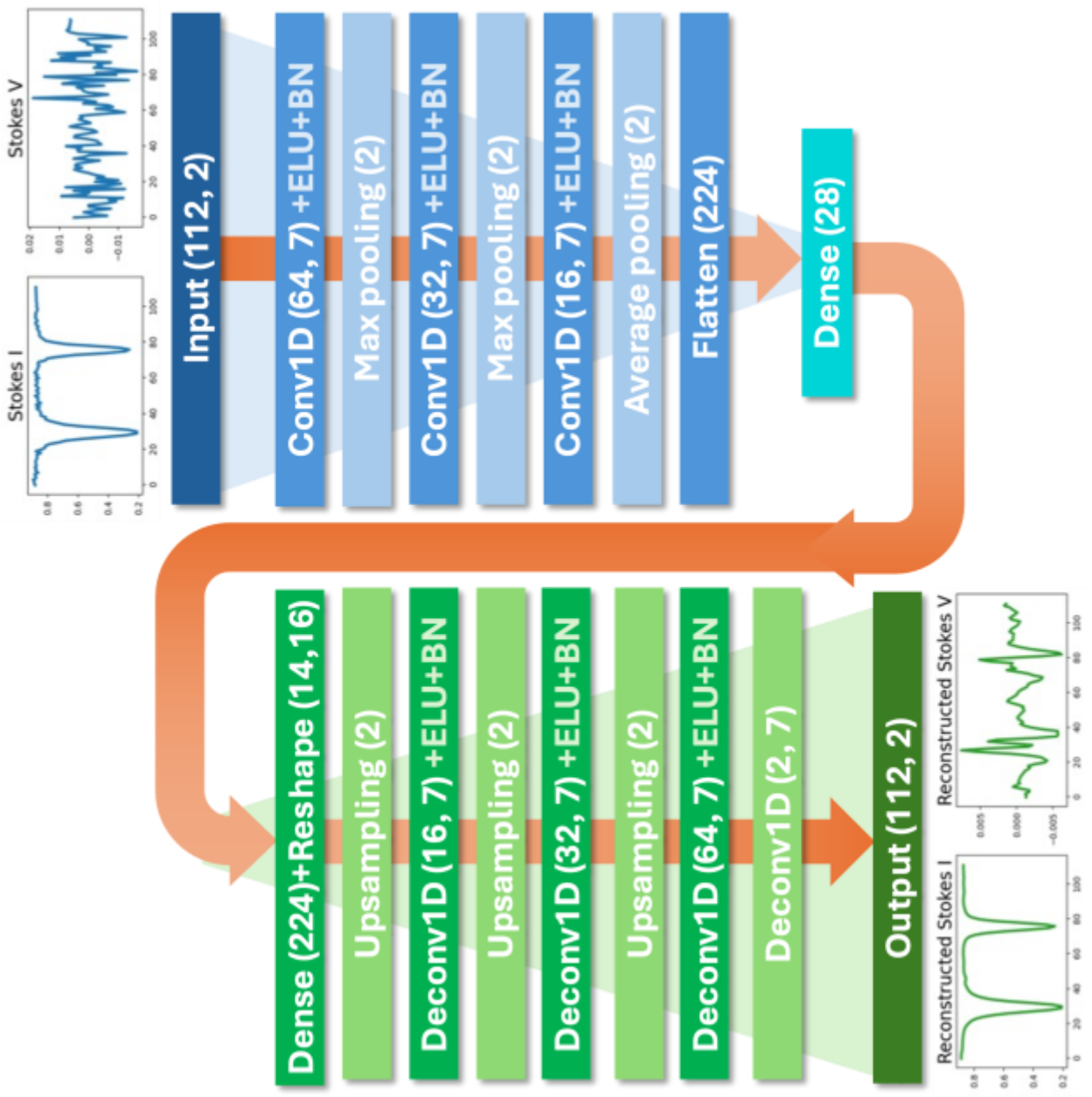

*Figure 1. Autoencoder architecture for spectral compression. The blue, cyan, and green shades represent the encoder, bottleneck, and decoder components, respectively. The layer/data shapes are shown in parentheses following each layer name.*

### 2.2. Dataset

Hinode/SP provides spectro-polarimetric (SP) map data, capturing spatial images of Stokes I, Q, U, and V for Fe I line pair profiles. Each pixel in the SP map contains intensity values for I, Q, U, and V across 112 wavelength points, specifically covering the line centers at 630.15 and 630.25 nm, as well as their continuums. This study focused on Stokes I and V profiles, representing the total intensity and the circular polarization, respectively. Therefore, the input data dimensions for the model are N×112×2 (the number of pixels × wavelength points × I and V parameters). Additionally, we incorporate temporal information from the data, allowing us to analyze spatio-temporal spectro-polarimetric data for a more comprehensive understanding.

We categorized our SP map data into two types: non-flare and pre-flare. Non-flare data consist of SP maps capturing ARs more than 24 hours prior to X-class flares, while pre-flare data encompass SP maps collected between 3 to 24 hours before X-class flares. In this context, non-flare maps are considered normal data, while pre-flare maps are treated as anomalous data.

For our analysis, we prepared eight non-flare SP maps for training and five for validating the compression model. Additionally, we collected four maps (three pre-flare maps and one during-flare map) for the X1.3 flare event in AR 13663 on May 5, 2024, for testing. All the datasets were downloaded from the Community Spectropolarimetric Analysis Center website [7,8]. Table 1 presents the SP map datasets used for model training, validation, and testing, labeled with their captured datetime and including details of the observation mode for a single FITS scan in each SP map.

*Table 1. SP maps datasets used for training, validation, and testing, including their observed datetime and observation mode from the first FITS scan of each SP map. The datetime format follows the convention of yyyymmdd_hhmmss, where the date is followed by the time.*

| Dataset | Observed datetime of SP maps | Observation mode | Pixel sampling scale in the dispersion direction [Ang/pix] | Pixel sampling scale in the slit direction [asec/pix] | Integration time [sec] | Step size of slit scanning |
|---|---|---|---|---|---|---|
| Training | 20160430_002505 | QT | 0.021549 | 0.317 | 1.6 | 1 |
| | 20201017_182230 | QT | 0.021549 | 0.317 | 1.6 | 1 |
| | 20201229_150006 | QT | 0.021549 | 0.317 | 1.6 | 1 |
| | 20210511_193050 (**SP3D20210511_193050.8C**) | QT | 0.021549 | **0.1585** | **4.8** | 1 |
| | 20210516_091405 | QT | 0.021549 | 0.317 | 1.6 | 1 |
| | 20210619_162335 | QT | 0.021549 | 0.317 | 1.6 | 1 |
| | 20210803_120106 (**SP3D20210803_120106.2C**) | QT | 0.021549 | **0.1585** | **4.8** | 1 |
| | 20211020_102805 | QT | 0.021549 | 0.317 | 1.6 | 1 |
| Validation | 20190515_115005 | QT | 0.021549 | 0.317 | 1.6 | 1 |
| | 20201107_112449 | QT | 0.021549 | 0.317 | 1.6 | 1 |
| | 20210312_080005 | QT | 0.021549 | 0.317 | 1.6 | 1 |
| | 20210618_224105 | QT | 0.021549 | 0.317 | 1.6 | 1 |
| | 20211201_010034 (**SP3D20211201_010034.5C**) | QT | 0.021549 | **0.1585** | **4.8** | 1 |
| Test | 20240504_143042 | QT | 0.021549 | 0.317 | 1.6 | 1 |
| | 20240505_020424 (**SP3D20240505_020424.2C**) | **FL** | 0.021549 | 0.317 | **0.8** | **2** |
| | 20240505_021351 | QT | 0.021549 | 0.317 | 1.6 | 1 |
| | 20240505_065322 (**SP3D20240505_065322.8C**) | **FL** | 0.021549 | 0.317 | **0.8** | **2** |

The observation mode parameters across the training, validation, and test sets were generally consistent across all the datasets. However, a few exceptions are highlighted in bold in Table 1, with their scan names listed in parentheses below the corresponding observed datetime. In the training and validation sets, the pixel sampling in the slit direction and integration time parameters were set to 0.1585 and 4.8, respectively, while these parameters were generally set to 0.317 and 1.6, respectively. For the test set, the observation mode, integration time, and scan step size parameters were set to ‘FL’, 0.8, and 2, respectively, unlike the ‘QT’, 1.6, and 1 setup used in all other cases. Despite these variations in observation modes, which result in differing noise levels across the data, the deep learning model is expected to perform robustly under these diverse conditions.

### 2.3. Training and evaluation

All the data preprocessing and the model training setups follow [6]. The SP maps used for training were resized to achieve a more focused field of view (FOV) on ARs. The total pixel counts for the eight resized SP maps used for training and the five SP maps used for validation were 911,860 and 875,940, respectively. We normalized the dataset for model training using min-max scaling for Stokes I and zero-mean scaling for Stokes V. The training process utilized a batch size of 512 over 1000 epochs, with the Adam optimizer and a reduced learning rate on the plateau parameter set with a patience of 50 epochs. Early stopping was also implemented with a patience of 100 epochs. The reconstruction loss function during training is defined as the sum of the mean absolute errors (MAEs) for the Stokes I and V profiles, as shown in Equation 1:

$$Loss_{recons} = MAE_I + MAE_V. \quad (1)$$

To evaluate the model's performance, similar to [6], we separately calculated the root mean square error (RMSE) between the original input profiles and the reconstructed output profiles of Stokes I and V spectra. The RMSE value calculated for each pixel in the SP maps serves as a measure of reconstruction accuracy, with pixels exhibiting higher RMSE values being more likely to be anomalous.

## 3. RESULTS

### 3.1. Model performance

Using the pre-trained compression model, we reconstructed the test spectra from three pre-flare SP maps and one during-flare SP map and computed the reconstruction error as RMSE for each pixel across all four SP maps. Table 2 presents the observational error ($\sigma_{obs}$), calculated as the mean standard deviation of the continuum within the wavelength pixel range from 0 to 15, along with the resulting mean reconstruction error ($\mu_{RMSE}$) values for both Stokes spectra across all four test SP maps.

*Table 2. Comparison of observational error and mean reconstruction error values.*

| SP map | $\sigma_{obs,I}$ | $\mu_{RMSE_I}$ | $\sigma_{obs,V}$ | $\mu_{RMSE_V}$ |
|---|---|---|---|---|
| 20240504_143042 | 0.0066 | 0.0060 | 0.0061 | 0.0056 |
| 20240505_020424 | 0.0072 | 0.0066 | 0.0083 | 0.0074 |
| 20240505_021351 | 0.0066 | 0.0060 | 0.0063 | 0.0058 |
| 20240505_065322 | 0.0073 | 0.0066 | 0.0079 | 0.0071 |

In all four cases, the mean RMSEs were lower than the observational errors for both Stokes parameters, indicating the model's satisfactory performance. As an example, Figure 2 presents the Stokes V spectra at four different spatial positions within AR 13663 (top-left plot), along with their reconstruction error ($RMSE_V$) values (top-right plot) and reconstructed profiles (bottom plot) in the SP map captured at 20240504_143042. The spectral positions are color-coded in the continuum image as follows: quiet Sun (red), pore (yellow), penumbra (magenta), and umbra (green), with these colors also representing their respective information in the other plots of Figure 2. The reconstructed profiles closely match the true Stokes V profiles, demonstrating a significant reduction in noise levels where applicable.

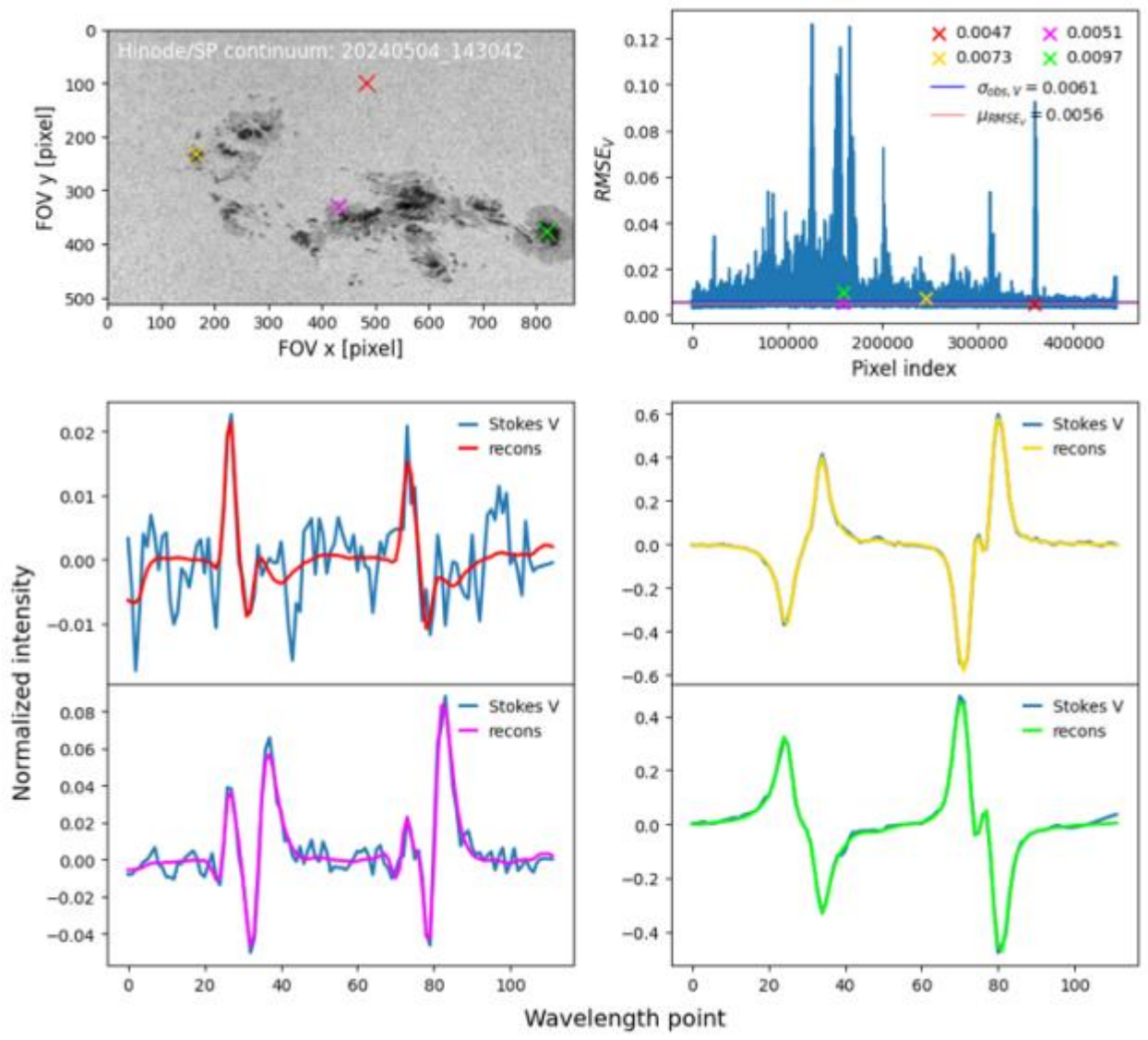

*Figure 2. Sample reconstructed profiles for different positions within AR 13663, including quiet Sun (red), pore (yellow), penumbra (magenta), and umbra (green). The top plots display the pixel positions on continuum image (left) and their reconstruction error values for Stokes V (right). The top-right plot also shows the Stokes V observational error ( $\sigma_{obs,V}$ ) and mean reconstruction error ( $\mu_{RMSE_V}$ ) levels. The bottom plot depicts the Stokes V profiles and their reconstructions for each selected pixel position.*

### 3.2. Anomalies with high reconstruction error

The Stokes V profile provides information about the photospheric line-of-sight (LOS) magnetic field component and represents the circular polarization of spectra caused by the Zeeman effect in magnetically sensitive lines. Using inversion or calibration techniques, LOS magnetograms—such as those generated by Hinode/SP or SDO/HMI [9,10]—are derived from raw Stokes V data. Therefore, examining the locations of detected anomalous Stokes V profiles on magnetograms s crucial to deduce the potential triggers of flares, driven by the buildup of magnetic energy.

Figure 3 illustrates spectral profiles that resulted in the highest reconstruction errors. Panels (a) to (c) display the pre-flare SP maps, while panel (d) corresponds to the during-flare SP map, with the time leading up to the start of the X1.3 flare indicated alongside their respective datetimes. Each panel displays an $RMSE_V$ heatmap and a separate overlaid magnetogram image in the left-side plots. The red cross on each magnetogram identifies the pixel with the highest $RMSE_V$ (h-$RMSE_V$) value. The spectral profiles and their corresponding reconstructions for these h-$RMSE_V$ pixels are shown in the right-side plots.

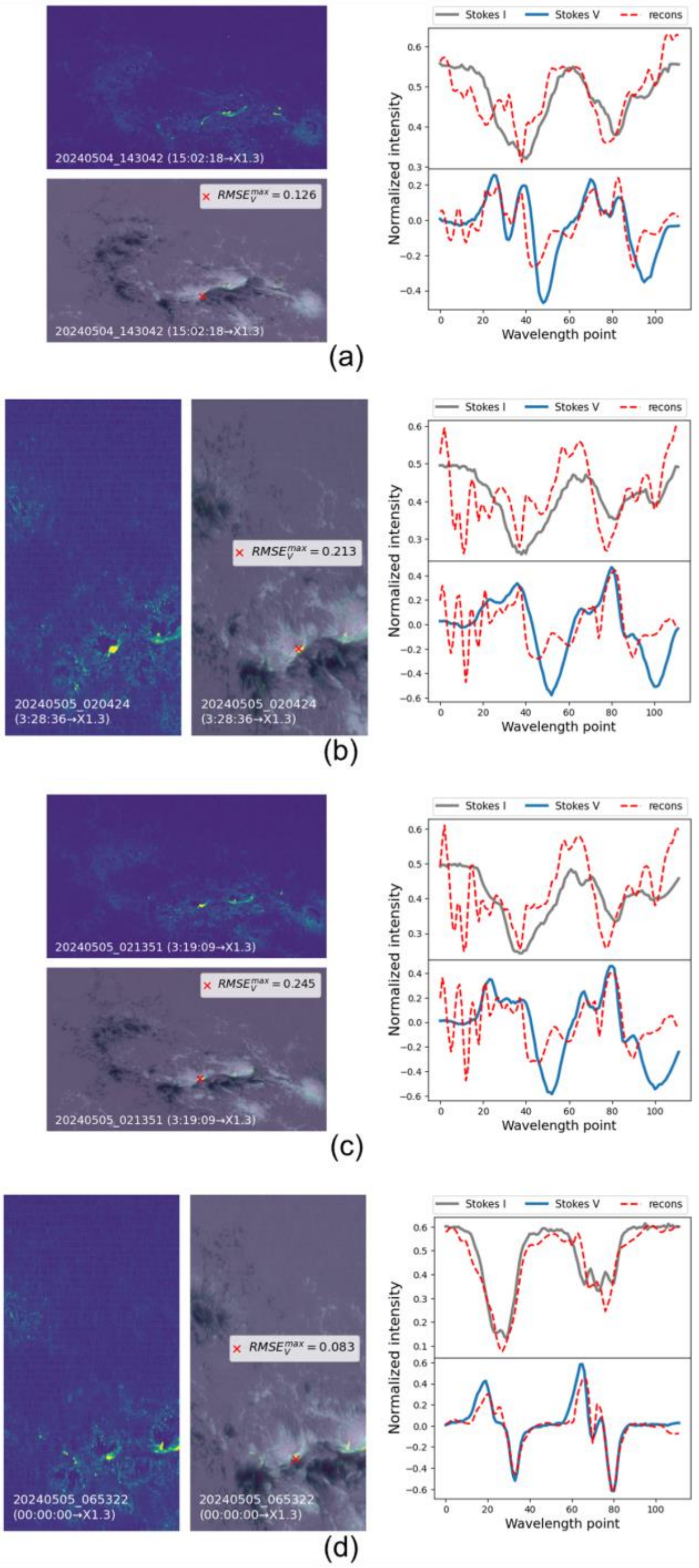

*Figure 3. Error heatmaps and detected anomalous spectral profiles. In each panel from (a) to (d), the left side shows an error heatmap and a magnetogram with the error heatmap overlaid, and the h-RMSE$_V$ pixel position marked by a red cross. The Stokes profiles and their reconstructions for the h-RMSE$_V$ pixel are presented on the right.*

In the heatmaps, darker areas represent lower reconstruction errors, while lighter areas indicate regions with higher $RMSE_V$ values. In all panels, the lighter regions in the heatmaps closely align with the polarity inversion lines (PIL) in the overlaid magnetogram. The spectral profiles of the h-$RMSE_V$ pixels in panels (a) to (c) exhibit unusual Stokes V shapes, with poor reconstructions. In contrast, panel (d) presents a stabilized Stokes V profile, showing improved reconstruction quality.

### 3.3. Flare starting region

We used image data from the Atmospheric Imaging Assembly [11] (AIA) 1600 Å filter and the Helioseismic and Magnetic Imager (HMI) on the Solar Dynamics Observatory (SDO) as references for the analysis. Bright regions on SDO/AIA 1600 images were used to identify potential flare-triggering sites for the X1.3 flare in AR 13663. Aligning these regions with SDO/HMI images enabled a comparison between the Hinode/SP and SDO/HMI magnetograms, allowing us to assess whether or not the locations of the detected anomalous Stokes V spectra correspond to the bright point locations.

The X1.3 flare began at 05:33, peaked at 05:59, and ended at 07:03 on May 5, 2024 [12]. The flare-triggering region is visually defined and marked positionally on the SDO images captured around the flare onset time to provide a reference for images captured at earlier timestamps, like those of the Hinode SP maps. Figure 4 presents two snapshots, taken just before and immediately after the flare onset, with red rectangles marking the potential flare-triggering regions on zoomed-in SDO/AIA 1600 images.

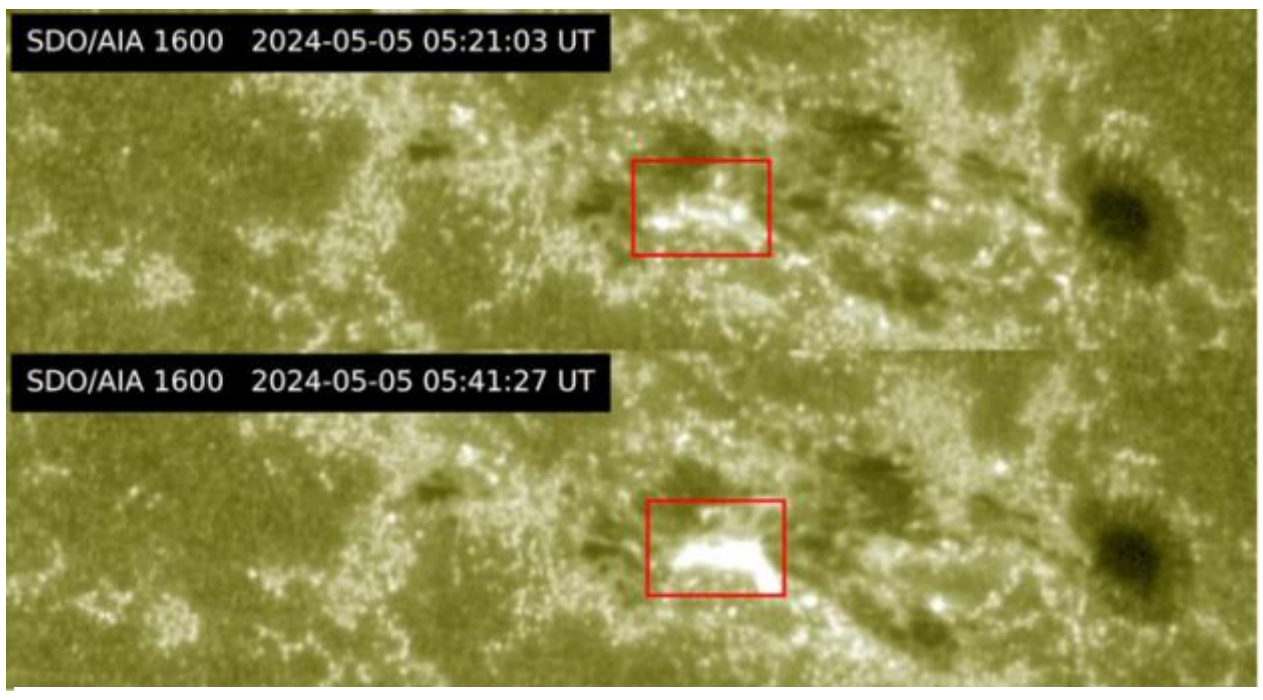

*Figure 4. SDO/AIA 1600 images showing the onset of the X1.3 flare event, with red rectangles highlighting potential flare-triggering regions. These regions are marked to serve as a reference for images captured at earlier timestamps, corresponding to the times of the Hinode/SP maps.*

To compare and align potential flare-triggering regions with our Hinode/SP map data, we prepared zoomed-in images from SDO/AIA 1600 and SDO/HMI, captured at timestamps closely matching those of the Hinode/SP maps (including three pre-flare maps and one during-flare map). Figure 5 displays zoomed views of AR 13663, showing SDO/AIA 1600 images (left column) and SDO/HMI magnetograms (right column).

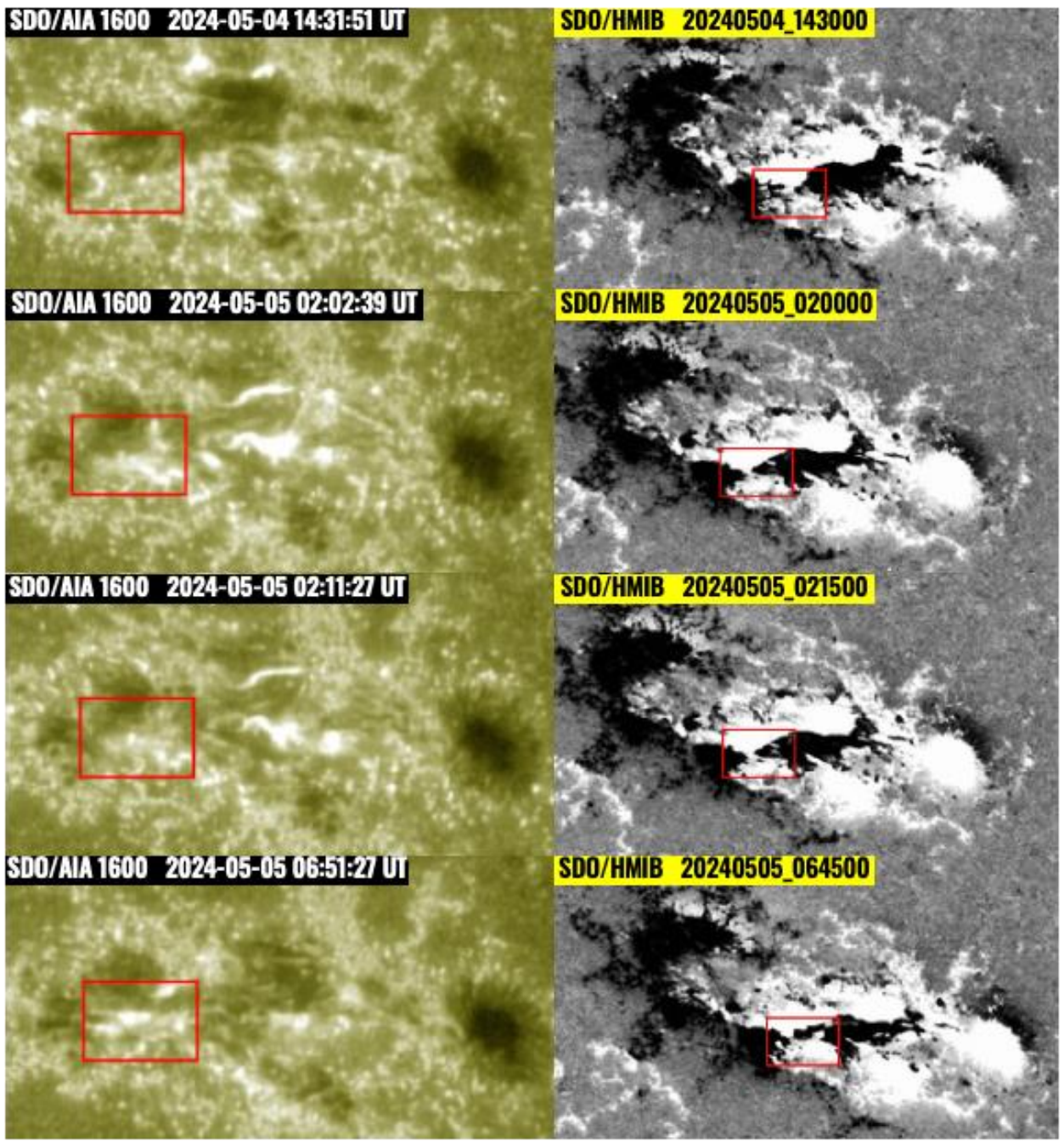

*Figure 4. Estimated flare triggering regions on SDO/AIA 1600 and SDO/HMIB images (zoomed views), captured at timestamps closely align with those of the Hinode/SP maps.*

The potential flare-triggering region, or region of interest (ROI), is marked with red rectangles based on visual approximation, corresponding to the highlighted area in Figure 4.

For a more detailed view, Figure 6 presents the zoomed-in ROI images indicated by red squares in Figure 5 (first and second columns), along with continuum images and contrast-enhanced magnetogram images from the Hinode/SP maps (third and fourth columns). These images enable a visual comparison with the SDO/HMI magnetograms. The red crosses on the Hinode/SP images correspond to the same h-RMSE$_V$ pixels highlighted in Figure 3, indicating that the anomalous spectra detected by the model are located within the ROI. Additionally, upon closely examining the locations of the red crosses on the Hinode/SP images and their corresponding positions on the SDO/AIA 1600 images, the latter exhibit varying levels of brightness, with the first and third images showing weaker brightness and with the others displaying stronger brightness.

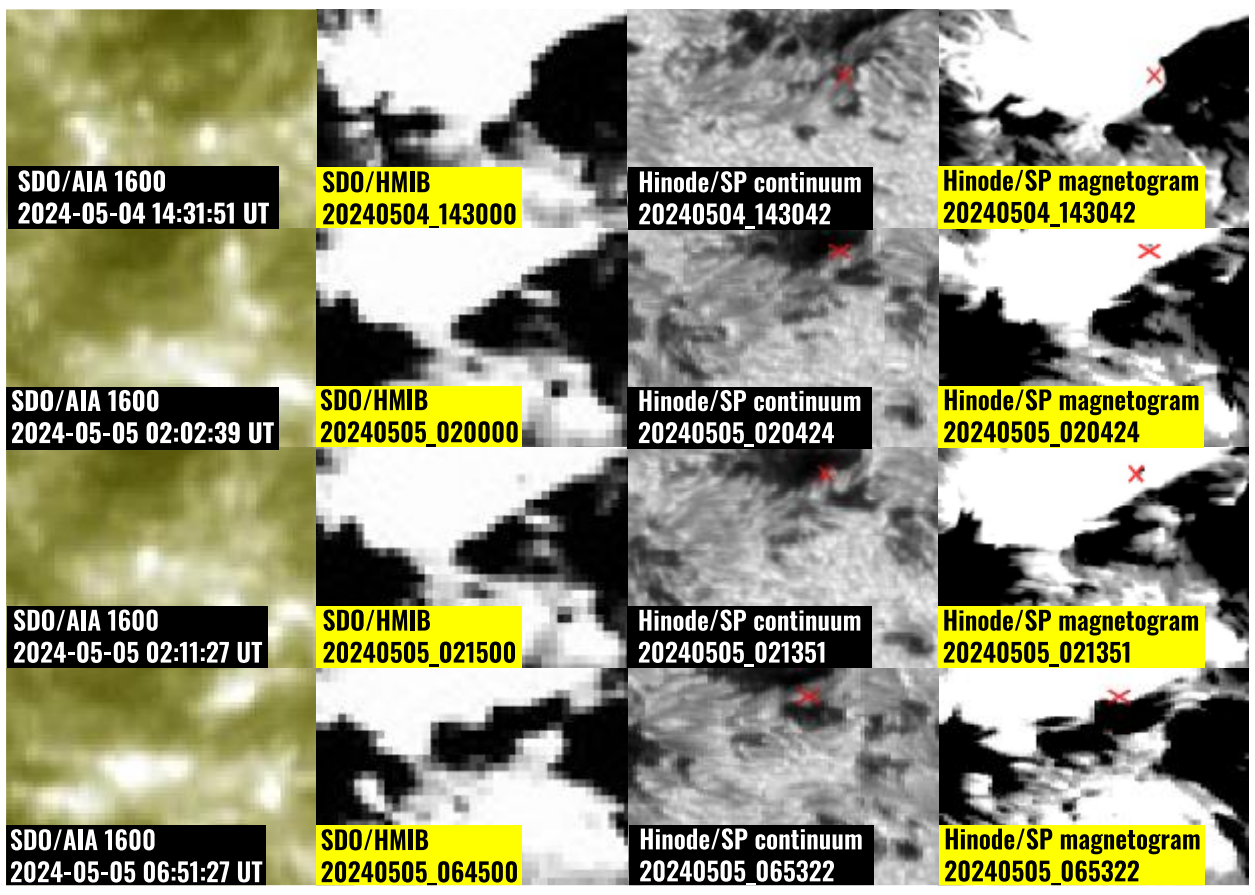

*Figure 6. Zoomed-in regions of interest in SDO and Hinode/SP maps. Red crosses indicate the h-RMSE$_V$ points on the SP maps.*

### 3.4. Anomalous focus areas in detail

To investigate the h-RMSE$_V$ points further, we analyzed the profiles of their neighboring pixels by zooming in on both the RMSE$_V$ heatmaps and the magnetograms, focusing on a 10 × 10-pixel area. Figure 7 presents the Stokes V spectral profiles of the surrounding pixels of the h-RMSE$_V$ points for a more detailed examination. Panels (a) to (c) show focused views of the h-RMSE$_V$ points from the pre-flare SP maps. In these heatmaps, the h-RMSE$_V$ points are distinctly localized, exhibiting notable differences in spectral shape and reconstruction errors compared with the surrounding pixels. In contrast, this trend is less pronounced in panel (d), the during-flare SP map, where the spectral profiles appear more generalized. This indicates that the change in pre-flare spectra diminishes and stabilizes after the flare onset.

## 4. DISCUSSION AND CONCLUSION

The spectra detected by the model exhibited atypical profiles, revealing interesting characteristics in the observations. The spectral shapes of the h-RMSE$_V$ pixels in the pre-flare SP maps from 20240504_143042, 20240505_020424, and 20240505_021351 showed asymmetries in the lobes, including double-peaked structures, unusual broadening of profiles, three-lobed shapes, and greater intensity imbalances among the lobes. These features indicate a complex magnetic field region, potentially influenced by a mix of supersonic downflows and upflows. The coexistence of these conditions, along with spatial proximity to PILs and their gradual appearance over time, creates a scenario conductive to magnetic reconnection or strong twists and shears in the magnetic fields. This suggests a clear signature of magnetic energy storage, which is a precursor to flare initiation [13,14].

The h-RMSE$_V$ pixels, particularly localized in the RMSE$_V$ heatmaps (Figure 7), exhibit significantly higher error values compared with their surrounding pixels. Identifying these pixels in magnetograms using current manual methods would present significant challenges, highlighting the effectiveness of deep learning for their detection. Additionally, the h-RMSE$_V$ pixels, calculated independently for each SP map, consistently appear in nearly the similar location (Figure 3). This consistency suggests that these pixels represent a time-evolving feature within the AR, further supporting the potential of this region as the site for magnetic energy buildup.

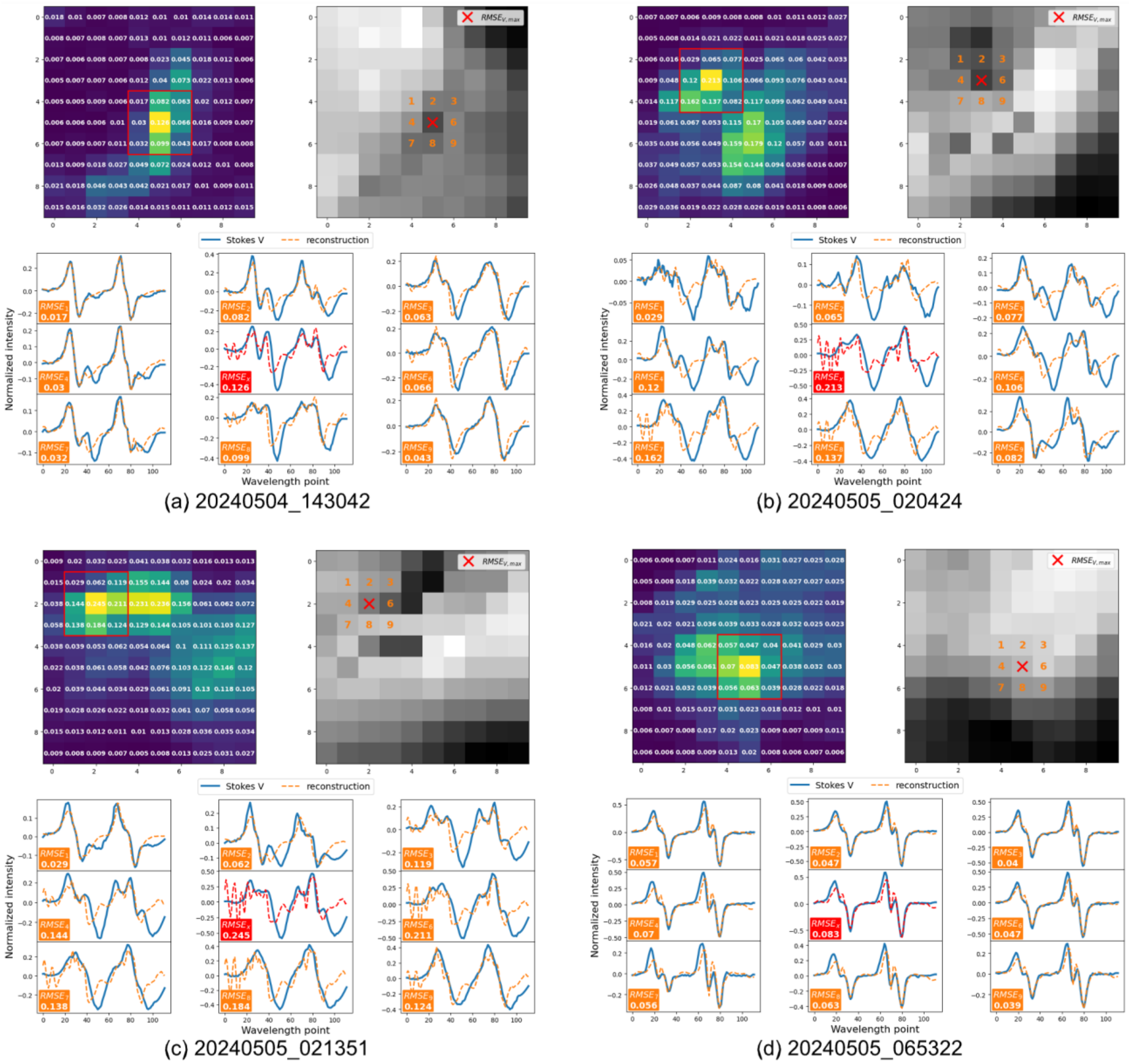

*Figure 5. Detailed views of anomalous regions. Each panel (a–d) shows a focused RMSE$_V$ heatmap (top-left) highlighting a 3×3-pixel grid outlined in red, with the h-RMSE$_V$ pixel (the brightest pixel) positioned at the center. This grid corresponds to the numbered pixels in orange on the matching focused magnetogram (top-right), where only the h-RMSE$_V$ pixel is marked with a red cross. The spectral profiles at the bottom display the profiles of the pixels within the 3×3 grid, identified by the orange indices (including the h-RMSE$_V$ pixel marked with the red cross). In the pre-flare panels (a–c), the brightest pixels with the highest error are highly localized in the RMSE$_V$ heatmaps, making it extremely difficult to detect in the magnetograms. In contrast, in the during-flare panel (d), the brightest pixel shows a less noticeable error value, and the localized trend is more dispersed.*

Regarding the reconstructed profiles of the h-RMSE$_V$ pixels (Figure 7), high fluctuations tend to appear on the blue side (0–20), indicating the model's difficulty in predicting extreme shapes of the left lobes in the left line cores. This aligns with our expectation because the approach is designed to detect unusual profiles as anomalies.

Unlike inversion techniques, which extract physical parameters from Stokes profiles to analyze the flare formation mechanism, our approach focuses on detecting and evaluating direct spectral shapes. This method accelerates the process of identifying flare-prone regions and provides a complementary tool for conventional approaches.

In conclusion, we developed a 1D-convolutional autoencoder-based compression model to detect anomalous Stokes V spectra from Hinode/SP data, with the goal of identifying potential flare-triggering points in the early stages before flare onset. Our results demonstrate that the model effectively identifies spectra with complex and unusual profiles in pre-flare spectro-polarimetric maps, collected 3 to 24 hours before flare initiation. The distinct shapes and locations of these spectra suggest that they are likely associated with magnetic energy storage and flare-prone regions. Our approach, which seeks to detect atypical spectral patterns as an early alert for upcoming flare events, holds promise for advancing solar spectral analysis and for supporting methods to predict solar flares. In future work, we aim to extend the detection method to apply it to the full Stokes spectra, enabling more sophisticated analysis and providing more reliable predictions.

## ACKNOWLEDGEMENTS

This work was supported by JST Grant Number JPMJFS2114, a university fellowship toward the creation of science technology innovation, and JST SPRING Grant Number JPMJSP2121.

Hinode is a Japanese mission developed and launched by ISAS/JAXA, collaborating with NAOJ as a domestic partner and with NASA and STFC (UK) as international partners. The scientific operation of the Hinode mission is being conducted by the Hinode science team organized at ISAS/JAXA. The team mainly consists of scientists from the institutes in partner countries. Support for the post-launch operation is provided by JAXA and NAOJ (Japan), STFC (U.K.), NASA, ESA, and NSC (Norway).

*1 XXXXXXXXXXXXXXX
(xxxxxxxxxxxxxxxxxxxxxxxxxxxxxxxxxxxxxxxxxx)

*2 VVVVVVVVVVVVVVVVVVVV
(vvvvvvvvvvvvvvvvvvvvvvvvvvvvvvvvvvv)